\pdfoutput=1

\documentclass[11pt]{article}

\usepackage{acl}

\usepackage{times}
\usepackage{latexsym}

\usepackage[T1]{fontenc}

\usepackage[utf8]{inputenc}

\usepackage{microtype}

\usepackage{bm}
\usepackage{amsmath}
\usepackage{graphicx}
\usepackage{amsfonts}
\DeclareMathOperator*{\E}{\mathbb{E}}

\usepackage{booktabs}
\usepackage{amssymb}

\title{Deliberation Networks and How to Train Them}

\author{Qingyun Dou \and Mark Gales \\
  University of Cambridge \\
  \texttt{\{qd212,mjfg100\}@cam.ac.uk} \\}

\begin{document}
\maketitle
\begin{abstract}
Deliberation networks are a family of sequence-to-sequence models, which have achieved state-of-the-art performance in a wide range of tasks such as machine translation and speech synthesis. A deliberation network consists of multiple standard sequence-to-sequence models, each one conditioned on the initial input and the output of the previous model.
During training, there are several key questions: whether to apply Monte Carlo approximation to the gradients or the loss, whether to train the standard models jointly or separately, whether to run an intermediate model in teacher forcing or free running mode, whether to apply task-specific techniques. Previous work on deliberation networks typically explores one or two training options for a specific task. This work introduces a unifying framework, covering various training options, and addresses the above questions. In general, it is simpler to approximate the gradients. When parallel training is essential, separate training should be adopted. Regardless of the task, the intermediate model should be in free running mode. For tasks where the output is continuous, a guided attention loss can be used to prevent degradation into a standard model.

\end{abstract}

\section{Introduction}

Auto-regressive sequence-to-sequence (seq2seq) models with attention mechanisms are used in a variety of areas including Neural Machine Translation (NMT) \cite{neubig2017neural,huang2016attention}, Automatic Speech Recognition (ASR) \cite{chan2015listen} and speech synthesis \cite{shen2018natural,wang2018style}, also known as Text-To-Speech (TTS). These models excel at connecting sequences of different length, but can be difficult to train. A standard approach is teacher forcing, which guides a model with reference output history during training. This makes the model unlikely to recover from its mistakes during inference, where the reference output is replaced by generated output. This issue is often referred to as exposure bias.
Several approaches have been introduced to tackle this issue, namely scheduled sampling \cite{bengio2015scheduled}, professor forcing \cite{lamb2016professor} and attention forcing \cite{dou2020attention, dou2021attention}. These approaches require sequential generation during training, and cannot be directly applied when parallel training is a priority.

Deliberation networks \cite{xia2017deliberation} are a family of multi-pass seq2seq models, and can be viewed as a parallelizable alternative approach to addressing exposure bias. Here the output sequence is generated in multiple passes, each one conditioned on the initial input and the output of the previous pass. For multi-pass seq2seq models, there are many choices to make during training, e.g. whether to update the parameters of each pass separately or jointly. Previous work \cite{xia2017deliberation, hu2020deliberation, dou2021deliberation} on deliberation networks typically focuses on a specific task, and explores one or two training options. This work introduces a unifying framework, covering various training options in a task-agnostic fashion, and then investigates task-specific techniques.



The novelties of this paper are as follows. First, section \ref{sec:chMP_framework} describes the framework from a probabilistic perspective, and section \ref{sec:chDelib_training} investigates a range of training approaches. In contrast, previous work \cite{xia2017deliberation, hu2020deliberation, hu2021transformer} takes a deterministic perspective, and describes the one or two training approaches adopted in the experiments. Second, section \ref{sec:chDelib_training} draws the connection between the training of deliberation networks and Minimum Bayes Risk (MBR) training. Leveraging the connection, the end of section \ref{sec:chDelib_training_joint} introduces a novel training approach, which approximates the loss, unlike previous work \cite{xia2017deliberation} approximating the gradients. The separate training approach described in section \ref{sec:chDelib_training_separate} is not novel, but its synergy with parallel training is pointed out for the first time. Finally, section \ref{sec:application} reviews several techniques facilitating the application of deliberation networks to specific tasks.

\section{Attention-based sequence-to-sequence generation}

Sequence-to-sequence (seq2seq) generation can be defined as the task of mapping an input sequence $\bm{x}_{1:L}$ to an output sequence $\bm{y}_{1:T}$ \cite{bengio2015scheduled}. The two sequences do not need to be aligned or have the same length. From a probabilistic perspective, a model $\bm{\theta}$ estimates the distribution of $\bm{y}_{1:T}$ given $\bm{x}_{1:L}$. For autoregressive models, this can be formulated as
\begin{equation}
\textstyle
p(\bm{y}_{1:T}|\bm{x}_{1:L}; \bm{\theta}) = \prod_{t=1}^{T} p(\bm{y}_{t} | \bm{y}_{1:t-1}, \bm{x}_{1:L}; \bm{\theta}) \label{eq:p_ar}
\end{equation}

\subsection{Encoder-attention-decoder architecture}

\begin{figure}
\centering
\includegraphics[width=7.8cm]{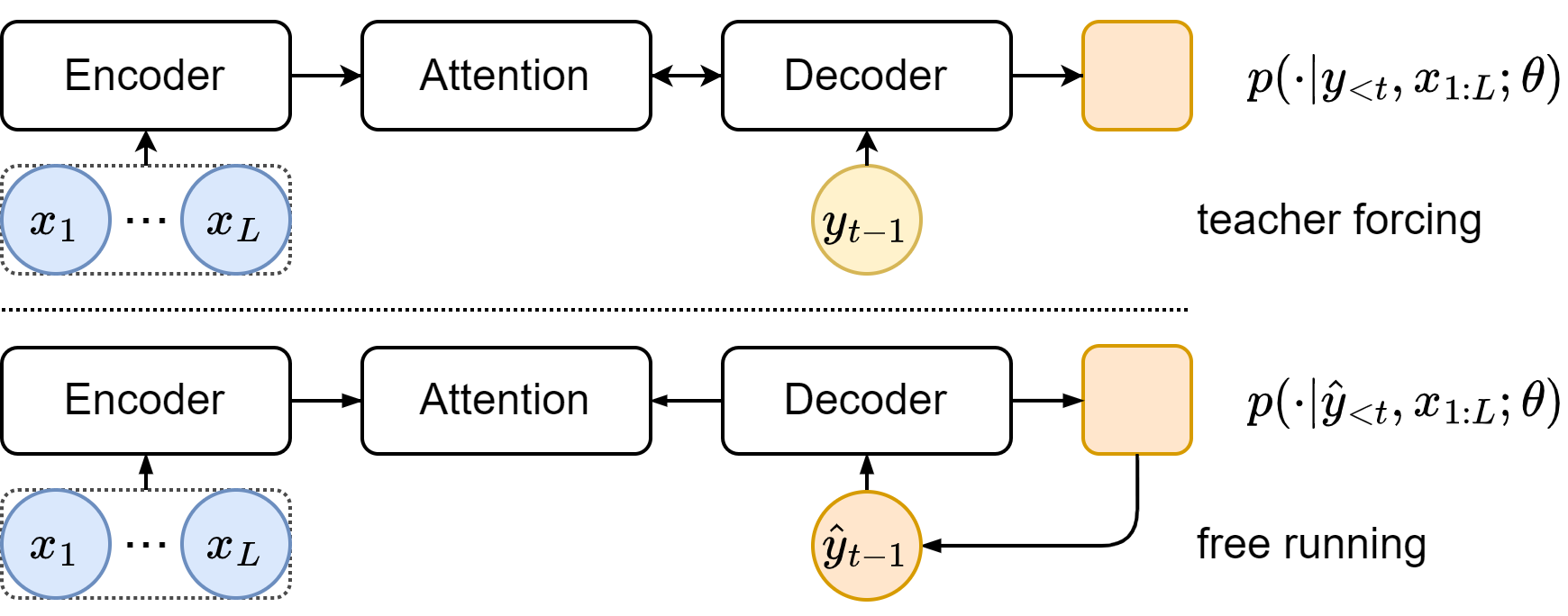}
\caption{A general attention-based encoder-decoder model, operating in teacher forcing mode; a circle depicts a token, and a rounded square depicts a distribution.}\label{fig:EAD}
\end{figure}

Attention-based sequence-to-sequence models usually have the encoder-attention-decoder architecture \cite{vaswani2017attention, lewis2020bart, tay2020efficient}. The distribution of a token is conditioned on the back-history, the input sequence and and attention map:
\begin{align}
p(\bm{y}_{t}|\bm{y}_{1:t-1}, \bm{x}_{1:L}; \bm{\theta}) &\approx p(\bm{y}_{t}|\bm{y}_{1:t-1}, \bm{\alpha}_{t}, \bm{x}_{1:L}; \bm{\theta}) \nonumber \\
&\approx p(\bm{y}_{t}|\bm{s}_{t}, \bm{c}_{t}; \bm{\theta}_{y}) \label{eq:py_general_p} 
\end{align}
where $\bm{\theta} = \{\bm{\theta}_{y}, \bm{\theta}_{s}, \bm{\theta}_{\alpha}, \bm{\theta}_{h}\}$; $\bm{\alpha}_{t}$ is an alignment vector, i.e. a set of attention weights; $\bm{s}_{t}$ is a state vector representing the output history $\bm{y}_{1:t-1}$, and $\bm{c}_{t}$ is a context vector summarizing $\bm{x}_{1:L}$ for time step $t$. Figure \ref{fig:EAD} shows a general encoder-attention-decoder model. The following equations give more details about how $\bm{\alpha}_{t}$, $\bm{s}_{t}$ and $\bm{c}_{t}$ can be computed:
\begin{gather*}
\bm{h}_{1:L} = f(\bm{x}_{1:L}; \bm{\theta}_{h}) \\
\bm{s}_{t} = f(\bm{y}_{1:t-1}; \bm{\theta}_{s}) \\
\textstyle \bm{\alpha}_{t} = f(\bm{s}_{t}, \bm{h}_{1:L}; \bm{\theta}_{\alpha}) \quad
\bm{c}_{t} = \sum_{l=1}^{L} \alpha_{t,l} \bm{h}_{l} \\
\hat{\bm{y}}_{t} \sim p(\cdot | \bm{s}_{t}, \bm{c}_{t}; \bm{\theta}_{y})
\end{gather*}
The encoder maps $\bm{x}_{1:L}$ to $\bm{h}_{1:L}$, considering information from the entire input sequence; $\bm{s}_{t}$ summarizes $\bm{y}_{1:t-1}$, considering only the past. With $\bm{h}_{1:L}$ and $\bm{s}_{t}$, the attention mechanism computes $\bm{\alpha}_{t}$, and then $\bm{c}_{t}$. Finally, the decoder estimates a distribution based on $\bm{s}_{t}$ and $\bm{c}_{t}$, and optionally generates an output token $\hat{\bm{y}}_{t}$.\footnote{ 
When computing the decoder state, the context vector can be optionally considered: $\bm{s}_{t} = f(\bm{y}_{1:t-1}, \bm{c}_{t-1}; \bm{\theta}_{s})$. For the discussions in this paper, it is not crucial whether the context vector is included.
}

\subsection{Inference and training} \label{sec:intro_inf_train}

During inference, given an input $\bm{x}_{1:L}$, the output $\hat{\bm{y}}_{1:T}$ can be obtained from the distribution estimated by the model $\bm{\theta}$:
\begin{gather}
\hat{\bm{y}}_{1:T} = \underset{\bm{y}_{1:T}}{\mathrm{argmax}} \, p( \bm{y}_{1:T} | \bm{x}_{1:L}; \bm{\theta})
\end{gather}
The exact search is often too expensive and is often approximated by greedy search for continuous output, or beam search for discrete output \cite{bengio2015scheduled}.

Conceptually, the model is trained to learn the natural distribution, e.g. through minimizing the KL-divergence between the natural distribution $p(\bm{y}_{1:T}|\bm{x}_{1:L})$ and the estimated distribution $p(\bm{y}_{1:T} | \bm{x}_{1:L}; \bm{\theta})$. In practice, this can be approximated by minimizing the Negative Log-Likelihood (NLL) over some training data $\{\bm{y}^{(n)}_{1:T}, \bm{x}^{(n)}_{1:L}\}_{1}^{N}$, sampled from the true distribution:
\begin{align}
\mathcal{L}(\bm{\theta}) &= \E_{\scalebox{0.8}{$\bm{x}_{1:L}$}} \mathrm{KL} \big(p(\bm{y}_{1:T}|\bm{x}_{1:L}) || p(\bm{y}_{1:T}|\bm{x}_{1:L}; \bm{\theta}) \big) \nonumber \\
&\propto - \textstyle \sum_{n=1}^{N} \log p(\bm{y}^{(n)}_{1:T}|\bm{x}^{(n)}_{1:L}; \bm{\theta}) \label{eq:pb_train}
\end{align}
$\mathcal{L}(\bm{\theta})$ denotes the loss. $N$ denotes the size of the training dataset; $n$ denotes the data index. To simplify the notation, the data index is omitted for the length of the sequences, although they also vary with the index. In the following sections, the sum over the training set $\sum_{n=1}^{N}$ will also be omitted.

For autoregressive models, and the sequence distribution $p(\bm{y}_{1:T}| \bm{x}_{1:L}; \bm{\theta})$ is factorized across time, as shown in equation \ref{eq:p_ar}. A key question then, is how to compute the token distribution $p(\bm{y}_{t}|\bm{y}_{1:t-1}, \bm{x}_{1:L}; \bm{\theta})$. For teacher forcing, at each time step $t$, the token distribution is computed with the correct output history $\bm{y}_{1:t-1}$. In this case, the loss can be written as:
\begin{equation} \label{eq:loss_y_iclr}
\begin{split}
    \mathcal{L}_{y}^{\tt T}(\bm{\theta}) &= -  \log p(\bm{y}_{1:T}| \bm{x}_{1:L}; \bm{\theta}) \\
    &= - \textstyle \sum_{t=1}^{T} \log p(\bm{y}_{t}| \bm{y}_{1:t-1}, \bm{x}_{1:L}; \bm{\theta})
\end{split}
\end{equation}

From a theoretical point of view, this approach yields the correct model (zero KL-divergence) if the following assumptions hold: 1) the model is powerful enough ; 2) the model is optimized correctly; 3) there is enough training data to approximate the expectation shown in equation \ref{eq:pb_train}. In practice, these assumptions are often not true, hence the model is prone to mistakes. From a different perspective, teacher forcing suffers from exposure bias \cite{ranzato2015sequence}, which refers to the following problem. During training, the model is guided by the reference output history. At the inference stage, however, the generated output history must be used. Hence there is a train-inference mismatch, and the errors accumulate along the inference process \cite{ranzato2015sequence}.

Many approaches have been introduced to tackle exposure bias, and there are mainly two lines of research. Scheduled sampling \cite{bengio2015scheduled, duckworth2019parallel} and professor forcing \cite{lamb2016professor} are prominent examples along the first line. These approaches guide a model with both the reference and the generated output history, and the goal is to learn the data distribution via maximizing the likelihood of the training data. To facilitate convergence, they often depend on a heuristic schedule or an auxiliary classifier, which can be difficult to design and tune \cite{bengio2015scheduled, guo2019new}. The second line is a series of sequence-level training approaches, leveraging reinforcement learning \cite{ranzato2015sequence}, minimum risk training \cite{shen2016minimum} or generative adversarial training \cite{yu2017seqgan}. Theses approaches guide a model with the generated output history. During training, the model operates in free running mode, and the goal is not to generate the reference output, but to optimize a sequence-level loss. However, many tasks do not have well established sequence-level objective metrics. Examples include speech synthesis, voice conversion, machine translation and text summarization \cite{tay2020efficient}. Both lines of research require generating output sequences, and this process is sequential for autoregressive models. In recent years, models based on the Transformer \cite{vaswani2017attention} have been widely used, and a key advantage is that when teacher forcing is used, training can be run in parallel across time. To efficiently generate output sequences from Transformer-based models, an approximation scheme \cite{duckworth2019parallel} has been proposed to parallelize scheduled sampling.

For the above training approaches, the model is trained at the token-level, i.e. the loss is computed for each token and summed across time. An alternative way of addressing exposure bias is to train the model at the sequence-level. Here the loss is computed for sequences instead of tokens, and the model sees not only the reference output during training. This type of approaches can be described in the framework of Minimum Bayes Risk (MBR) training. Assume that there is a distance metric $\mathcal{D} (\bm{y}_{1:T}, \underline{\bm{y}}_{1:\underline{T}})$ between the reference output $\bm{y}_{1:T}$ and a random output $\underline{\bm{y}}_{1:\underline{T}}$, whose probability is estimated with the model $\bm{\theta}$. $\mathcal{D}$ is minimal when the two sequences are equal. MBR training minimizes its expected value:
\begin{equation} \label{eq:loss_y_bayes}
\begin{split}
    \mathcal{L}_{y}^{\tt B}(\bm{\theta}) &=  \sum_{\underline{\bm{y}}_{1:\underline{T}} \in \mathcal{Y}}  p(\underline{\bm{y}}_{1:\underline{T}} | \bm{x}_{1:L}; \bm{\theta}) \mathcal{D} (\bm{y}_{1:T}, \underline{\bm{y}}_{1:\underline{T}})
\end{split}
\end{equation}
where $\mathcal{Y}$ is the entire output space.

\section{Network Architecture} \label{sec:chMP_framework}


Deliberation networks are inspired by a common human behavior: when producing a sequence, be it text or speech, we often revise the initial output to improve its quality. For example, to write a good article, we usually first create a draft and then polish it. To record a section of an audio book, the readers often record several times until the quality is good enough.

\begin{figure}
\centering
\includegraphics[width=7.8cm]{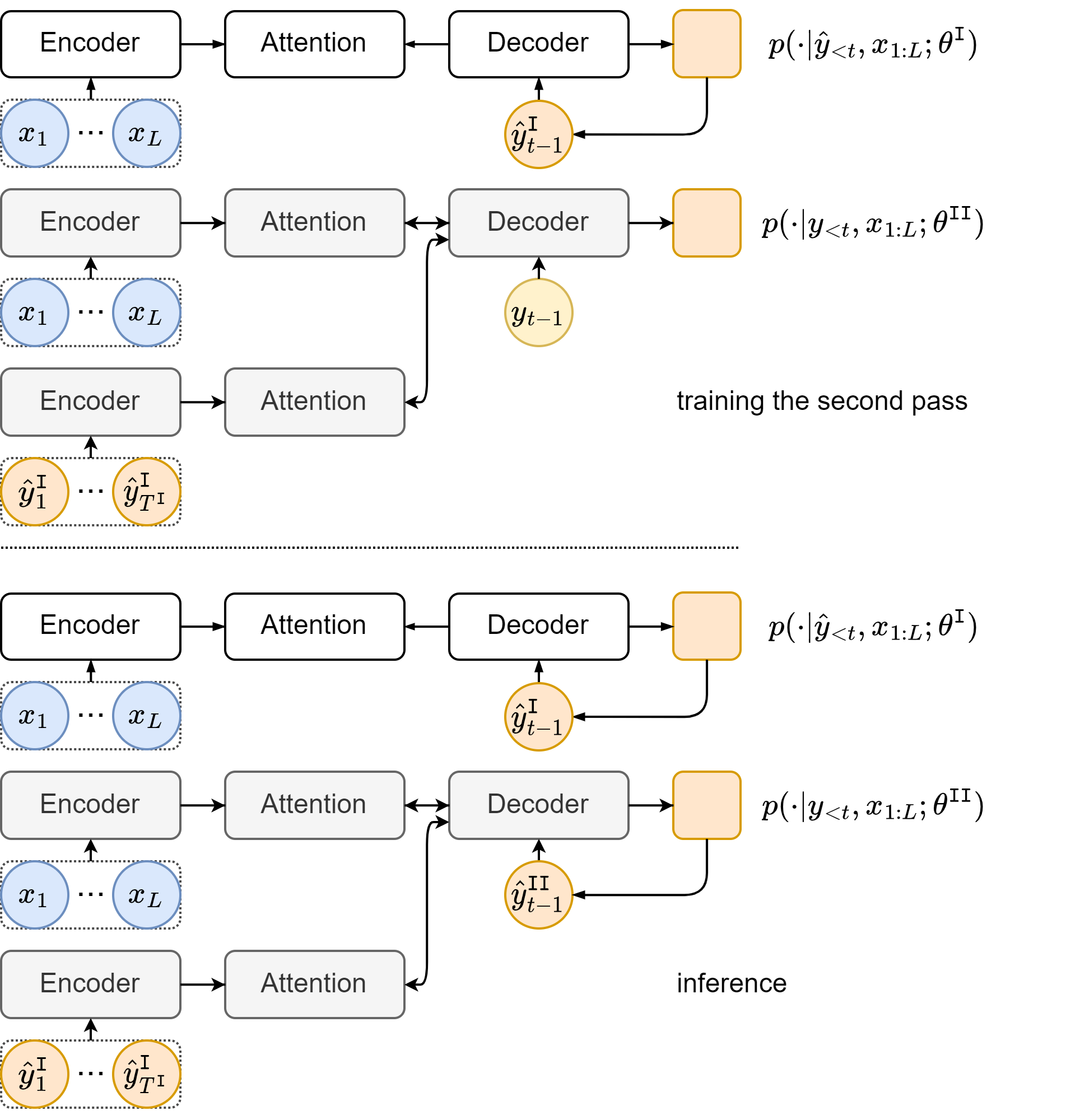}
\caption{A two-pass deliberation network; the clear blocks depict the first pass; the shaded blocks depict the second pass.} \label{fig:MP}
\end{figure}

A deliberation network consists of multiple models. Its output is generated in multiple passes, each one conditioned on the initial input and the previous free running output. With the iterative refinement, the final output is expected to be better than the previous ones. For deliberation networks, an essential element is to condition all but the first model on its previous free running output. This allows the later models to learn to correct the free running output, alleviating exposure bias. Without loss of generality, this section describes a two-pass deliberation network, shown in figure \ref{fig:MP}.

In terms of notation, $\bm{x}_{1:L}$ and $\bm{y}_{1:T}$ denote the input and reference output; $\bm{\theta}^{\tt I}$ and $\bm{\theta}^{\tt II}$ denote the first-pass and second-pass models; $\bm{y}^{\tt I}_{1:T^{\tt I}}$ denotes an intermediate output sequence. The deliberation network models $p(\bm{y}_{1:T} | \bm{x}_{1:L})$ as
\begin{align} \label{eq:prob_mp}
&p(\bm{y}_{1:T} | \bm{x}_{1:L}; \bm{\theta}^{\tt I}, \bm{\theta}^{\tt II}) = \textstyle \sum_{\bm{y}^{\tt I}_{1:T^{\tt I}} \in \mathcal{Y}} \\
&p(\bm{y}^{\tt I}_{1:T^{\tt I}} | \bm{x}_{1:L}; \bm{\theta}^{\tt I})
p(\bm{y}_{1:T} | \bm{y}^{\tt I}_{1:T^{\tt I}}, \bm{x}_{1:L}; \bm{\theta}^{\tt II}) \nonumber
\end{align}
if $\bm{y}^{\tt I}_{1:T^{\tt I}}$ is discrete, and
\begin{align}
&p(\bm{y}_{1:T} | \bm{x}_{1:L}; \bm{\theta}^{\tt I}, \bm{\theta}^{\tt II}) = 
\textstyle \int_{\bm{y}^{\tt I}_{1:T^{\tt I}} \in \mathcal{Y}} \\
&p(\bm{y}^{\tt I}_{1:T^{\tt I}} | \bm{x}_{1:L}; \bm{\theta}^{\tt I})
p(\bm{y}_{1:T} | \bm{y}^{\tt I}_{1:T^{\tt I}}, \bm{x}_{1:L}; \bm{\theta}^{\tt II}) d\bm{y}^{\tt I}_{1:T^{\tt I}} \nonumber
\end{align}
if $\bm{y}^{\tt I}_{1:T^{\tt I}}$ is continuous. The summation/integration is over $\mathcal{Y}$, the entire space of $\bm{y}^{\tt I}_{1:T^{\tt I}}$. $p(\bm{y}^{\tt I}_{1:T^{\tt I}} | \bm{x}_{1:L}; \bm{\theta}^{\tt I})$ is computed by $\bm{\theta}^{\tt I}$, a standard single-pass model. $p(\bm{y}_{1:T} | \bm{y}^{\tt I}_{1:T^{\tt I}}, \bm{x}_{1:L}; \bm{\theta}^{\tt II})$ is computed by $\bm{\theta}^{\tt II}$, a model with an additional attention mechanism over $\bm{y}^{\tt I}_{1:T^{\tt I}}$. $\bm{\theta}^{\tt I}$ and $\bm{\theta}^{\tt II}$ have different time steps. At time $t$, assuming $\bm{y}^{\tt I}_{1:T^{\tt I}}$ is available, $\bm{\theta}^{\tt II}$ operates as follows.
\begin{gather}
\bm{s}_{t} = f(\bm{y}_{1:t-1}; \bm{\theta}_{s}^{\tt II}) \label{eq:mp_s} \\
\begin{split}
\bm{h}_{x, 1:L} = f(\bm{x}_{1:L}; \bm{\theta}_{h,x}^{\tt II}) \\
\bm{h}_{y, 1:T^{\tt I}} = f(\bm{y}^{\tt I}_{1:T^{\tt I}}; \bm{\theta}_{h,y}^{\tt II})
\end{split} \nonumber \\
\begin{split}
\bm{\alpha}_{x,t} = f(\bm{s}_{t}, \bm{h}_{x,1:L}; \bm{\theta}_{\alpha,x}^{\tt II}) \quad 
\bm{c}_{x,t} =\textstyle \sum_{l=1}^{L} \alpha_{x,t,l} \bm{h}_{x,l} \\
\bm{\alpha}_{y,t} = f(\bm{s}_{t}, \bm{h}_{y,1:T^{\tt I}}; \bm{\theta}_{\alpha,y}^{\tt II}) \quad 
\bm{c}_{y,t} =\textstyle \sum_{l=1}^{T^{\tt I}} \alpha_{y,t,l} \bm{h}_{y,l}
\end{split} \nonumber \\
\hat{\bm{y}}_{t}^{\tt II} \sim p(\cdot | \bm{s}_{t}, \bm{c}_{x,t}, \bm{c}_{y,t}; \bm{\theta}_{y}^{\tt II}) \nonumber
\end{gather}

$\bm{\theta}^{\tt II}$ is built upon $\bm{\theta}^{\tt I}$, and has an additional encoder-attention pair. One pair $\{ \bm{\theta}_{h,x}^{\tt II}, \bm{\theta}_{\alpha,x}^{\tt II} \}$ is for the initial input $\bm{x}_{1:L}$; the additional pair $\{ \bm{\theta}_{h,y}^{\tt II}, \bm{\theta}_{\alpha,y}^{\tt II} \}$ is for the intermediate output $\bm{y}^{\tt I}_{1:T^{\tt I}}$. The encoder for $\bm{x}_{1:L}$ shares the same parameters as that of $\bm{\theta}^{\tt I}$, i.e. $\bm{\theta}_{h,x}^{\tt II} = \bm{\theta}_{h}^{\tt I}$. The probability of $\bm{y}_{t}$ depends on $\bm{s}_{t}$, $\bm{c}_{x,t}$ and $\bm{c}_{y,t}$. $\bm{s}_{t}$ is the state vector tracking the output history, and is used by both attention mechanisms. $\bm{c}_{x,t}$ and $\bm{c}_{y,t}$ summarize $\bm{x}_{1:L}$ and $\bm{y}^{\tt I}_{1:T^{\tt I}}$ respectively. The intermediate output of $\bm{\theta}^{\tt I}$, such as $\bm{s}^{\tt I}_{1:T^{\tt I}}$ and $\bm{c}^{\tt I}_{1:T^{\tt I}}$, can be combined with the $\bm{y}^{\tt I}_{1:T^{\tt I}}$ as the input to $\bm{\theta}_{h,y}^{\tt II}$.

During inference, the generated output history replaces the reference, and equation \ref{eq:mp_s} becomes $\bm{s}_{t} = f(\hat{\bm{y}}_{1:t-1}^{\tt II}; \bm{\theta}_{s}^{\tt II})$. The decoding of $\bm{\theta}^{\tt II}$ begins when that of $\bm{\theta}^{\tt I}$ is complete.

By default, the rest of this paper assumes that $\bm{y}^{\tt I}_{1:T^{\tt I}}$ is discrete, and most discussions are agnostic to the continuity of $\bm{y}^{\tt I}_{1:T^{\tt I}}$. When the continuity does make a difference, the discrete and continuous cases will be discussed separately.

\section{Training} \label{sec:chDelib_training}

\subsection{Joint Training} \label{sec:chDelib_training_joint}



\begin{table}
\centering
\caption{\label{tab:abb_math} Abbreviated expressions; these terms appear repeatedly in the equations, and are introduced to improve readability.}
\begin{tabular}{@{}l|l@{}}
\toprule
$F^{\tt I}$ & $p(\bm{y}^{\tt I}_{1:T^{\tt I}} | \bm{x}_{1:L}; \bm{\theta}^{\tt I})$                                      \\ \midrule
$F^{\tt II}$        & $p(\bm{y}_{1:T} | \bm{y}^{\tt I}_{1:T^{\tt I}}, \bm{x}_{1:L}; \bm{\theta}^{\tt II})$    \\ \midrule
$\hat{F}^{{\tt I} (m)}$  & $p(\hat{\bm{y}}^{{\tt I} (m)}_{1:T^{{\tt I} (m)}} | \bm{x}_{1:L}; \bm{\theta}^{\tt I})$  \\ \midrule
$\hat{F}^{{\tt II} (m)}$      & $p(\bm{y}_{1:T} | \hat{\bm{y}}^{{\tt I} (m)}_{1:T^{{\tt I} (m)}}, \bm{x}_{1:L}; \bm{\theta}^{\tt II})$             \\ \bottomrule
\end{tabular}
\end{table}

In theory, ${\bm\theta}^{\tt I}$ and ${\bm\theta}^{\tt II}$ can be trained by directly maximizing the log of the likelihood in equation \ref{eq:prob_mp}, and the loss function $\check{\mathcal{L}}_{y}$ is
\begin{align} 
\check{\mathcal{L}}_{y}(\bm{\theta}^{\tt I}, \bm{\theta}^{\tt II}) &= -\log p(\bm{y}_{1:T} | \bm{x}_{1:L}; \bm{\theta}^{\tt I}, \bm{\theta}^{\tt II}) \label{eq:mp_loss_naive} \\
&= -\log \textstyle \sum_{\bm{y}^{\tt I}_{1:T^{\tt I}} \in \mathcal{Y}} F^{\tt I} F^{\tt II} \nonumber
\end{align}
$F^{\tt I}$ and $F^{\tt II}$ are abbreviated expressions defined in table \ref{tab:abb_math}. As discussed in section \ref{sec:intro_inf_train}, if $\check{\mathcal{L}}_{y}$ were fully optimized, the KL-divergence between the model distribution $p(\bm{y}_{1:T} | \bm{x}_{1:L}; \bm{\theta}^{\tt I}, \bm{\theta}^{\tt II})$ and the true distribution $p(\bm{y}_{1:T} | \bm{x}_{1:L})$ would be zero. In general, the actual divergence is limited by the model, data and training. In particular, for deliberation networks, the sum over $\bm{y}^{\tt I}_{1:T^{\tt I}}$ is intractable due to the prohibitively large space of $\bm{y}^{\tt I}_{1:T^{\tt I}}$. A Monte Carlo estimator is often used to approximate either the loss or the gradients. The gradients of $\check{\mathcal{L}}_{y}$ w.r.t the model parameters $\bm{\theta}^{\tt I}$ and $\bm{\theta}^{\tt II}$ are
\begin{align}
&\triangledown_{\bm{\theta}^{\tt I}} \check{\mathcal{L}}_{y}(\bm{\theta}^{\tt I}, \bm{\theta}^{\tt II})
= -\displaystyle \frac{ { \sum_{\bm{y}^{\tt I}_{1:T^{\tt I}} \in \mathcal{Y}}} F^{\tt II} (\triangledown_{\bm{\theta}^{\tt I}} F^{\tt I}) }
{ \sum_{\bm{y}^{\tt I}_{1:T^{\tt I}} \in \mathcal{Y}} F^{\tt I} F^{\tt II} } \label{eq:mp_loss_naive_grad1} \\
&\triangledown_{\bm{\theta}^{\tt II}} \check{\mathcal{L}}_{y}(\bm{\theta}^{\tt I}, \bm{\theta}^{\tt II}) = -\displaystyle \frac{ { \sum_{\bm{y}^{\tt I}_{1:T^{\tt I}} \in \mathcal{Y}}} F^{\tt I} (\triangledown_{\bm{\theta}^{\tt II}} F^{\tt II}) }
{ \sum_{\bm{y}^{\tt I}_{1:T^{\tt I}} \in \mathcal{Y}} F^{\tt I} F^{\tt II} } \label{eq:mp_loss_naive_grad2} 
\end{align}
The sum over the output space appears twice in equations \ref{eq:mp_loss_naive_grad1} and \ref{eq:mp_loss_naive_grad2}, which makes the gradient computation more difficult than necessary. A commonly used technique is to instead minimize an upper bound $\mathcal{L}_{y}$ \cite{xia2017deliberation}:
\begin{align}
\mathcal{L}_{y} (\bm{\theta}^{\tt I}, \bm{\theta}^{\tt II}) 
&= - \textstyle \sum_{\bm{y}^{\tt I}_{1:T^{\tt I}} \in \mathcal{Y}} F^{\tt I}
\log F^{\tt II}\label{eq:mp_loss_lb} \\
&\geq \check{\mathcal{L}}_{y} (\bm{\theta}^{\tt I}, \bm{\theta}^{\tt II}) \nonumber
\end{align}

The upper bound is derived with the concavity of the log function and the fact that $\sum_{\bm{y}^{\tt I}_{1:T^{\tt I}} \in \mathcal{Y}} p(\bm{y}^{\tt I}_{1:T^{\tt I}} | \bm{x}_{1:L}; \bm{\theta}^{\tt I}) = 1$. The gradients of $\mathcal{L}_{y}$ w.r.t. $\bm{\theta}^{\tt I}$ and $\bm{\theta}^{\tt II}$ are
\begin{align}
&\triangledown_{\bm{\theta}^{\tt I}} \mathcal{L}_{y}(\bm{\theta}^{\tt I}, \bm{\theta}^{\tt II}) = \label{eq:mp_loss_grad1} \\
&- \textstyle \sum_{\bm{y}^{\tt I}_{1:T^{\tt I}} \in \mathcal{Y}} F^{\tt I} (\log F^{\tt II}) (\triangledown_{\bm{\theta}^{\tt I}} \log F^{\tt I}) \nonumber \\
&\triangledown_{\bm{\theta}^{\tt II}} \mathcal{L}_{y}(\bm{\theta}^{\tt I}, \bm{\theta}^{\tt II}) = \label{eq:mp_loss_grad2} \\
& - \textstyle \sum_{\bm{y}^{\tt I}_{1:T^{\tt I}} \in \mathcal{Y}} F^{\tt I} (\triangledown_{\bm{\theta}^{\tt II}} \log F^{\tt II}) \nonumber 
\end{align}

Equation \ref{eq:mp_loss_grad1} is derived using the identity $\triangledown_{\bm{\theta}} f(\bm{\theta}) = f(\bm{\theta}) \triangledown_{\bm{\theta}} \log f(\bm{\theta})$. Compared with the previous gradients shown in equations \ref{eq:mp_loss_naive_grad1} and \ref{eq:mp_loss_naive_grad2}, the new gradients, $\triangledown_{\bm{\theta}^{\tt I}} \mathcal{L}_{y}$ and $\triangledown_{\bm{\theta}^{\tt II}} \mathcal{L}_{y}$, are simpler in the sense that the summation over the output space appears only once.

Comparing equations \ref{eq:mp_loss_lb} and \ref{eq:loss_y_bayes}, it can be seen that for $\bm{\theta}^{\tt I}$, the loss fits into the framework of Minimum Bayes Risk (MBR) training, described in section \ref{sec:intro_inf_train}. Here the risk is defined as $ - \log p(\bm{y}_{1:T} | \bm{y}^{{\tt I}}_{1:T^{{\tt I}}}, \bm{x}_{1:L}; \bm{\theta}^{\tt II})$. The following subsections will describe two ways of using Monte Carlo approximation, in order to approximate the summation over the output space. They differ in whether to approximate the loss or the gradient, but they can adopt the same sampling process.

\subsubsection{Approximating Gradients}

Applying Monte Carlo approximation, the gradients $\triangledown_{\bm{\theta}^{\tt I}} \mathcal{L}_{y}$ and $\triangledown_{\bm{\theta}^{\tt II}} \mathcal{L}_{y}$ are estimated as
\begin{align}
& \triangledown_{\bm{\theta}^{\tt I}} \mathcal{L}_{y}(\bm{\theta}^{\tt I}, \bm{\theta}^{\tt II}) \approx \label{eq:mp_loss_grad1_approx} \\
&- \textstyle \frac{1}{M} \sum_{m=1}^{M} (\log \hat{F}^{{\tt II} (m)}) (\triangledown_{\bm{\theta}^{\tt I}} \log \hat{F}^{{\tt I} (m)}) \nonumber \\
& \triangledown_{\bm{\theta}^{\tt II}} \mathcal{L}_{y}(\bm{\theta}^{\tt I}, \bm{\theta}^{\tt II}) \approx \label{eq:mp_loss_grad2_approx} \\
&- \textstyle \frac{1}{M} \sum_{m=1}^{M} \triangledown_{\bm{\theta}^{\tt II}} \log \hat{F}^{{\tt II} (m)} \nonumber 
\end{align}

\noindent $\hat{F}^{{\tt I} (m)}$ and $\hat{F}^{{\tt II} (m)}$ are abbreviated expressions defined in table \ref{tab:abb_math}.
$\{ \hat{\bm{y}}^{{\tt I} (1)}_{1:T^{{\tt I} (1)}}, ..., \hat{\bm{y}}^{{\tt I} (M)}_{1:T^{{\tt I} (M)}} \}$ are $M$ i.i.d. samples drawn from the distribution $p(\bm{y}^{\tt I}_{1:T^{\tt I}} | \bm{x}_{1:L}; \bm{\theta}^{\tt I})$. The Monte Carlo estimator is unbiased, and its variance is proportional to $\frac{1}{M}$. There is a trade-off between the variance and the computational cost: fewer samples results in higher variance, but lower cost. For SGD-based optimization, noisy estimates of the gradients are used, and this variance adds another level of noise.

The sampling process is often realized by beam search or (a noisy version of) greedy search \cite{prabhavalkar2018minimum}, as described in section \ref{sec:intro_inf_train}. Once the sampling process is complete, $\triangledown_{\bm{\theta}^{\tt I}} \mathcal{L}_{y}$ and $\triangledown_{\bm{\theta}^{\tt II}} \mathcal{L}_{y}$ can be computed. For $\bm{\theta}^{\tt I}$, the training is equivalent to MBR training.
For $\bm{\theta}^{\tt II}$, the training can be viewed as teacher forcing, because to compute the conditional probability of the reference output, the reference back-history is used.

\subsubsection{Approximating Loss}

Alternatively, Monte Carlo approximation can be applied to the loss $\mathcal{L}_{y}$ shown in equation \ref{eq:mp_loss_lb}, before deriving the gradients. Let $\bar{\mathcal{L}}_{y}$\footnote{
After approximating the loss, a bar is added to the symbols, in order to facilitate comparison with the case where Monte Carlo approximation is applied to the gradients.
}
denote the approximation:
\begin{align}
&\mathcal{L}_{y}(\bm{\theta}^{\tt I}, \bm{\theta}^{\tt II}) \approx \bar{\mathcal{L}}_{y}(\bm{\theta}^{\tt I}, \bm{\theta}^{\tt II}) \label{eq:mp_loss_approx} \\
&= \textstyle - \frac{1}{M} \sum_{m=1}^{M} \log p(\bm{y}_{1:T} | \hat{\bm{y}}^{{\tt I} (m)}_{1:T^{{\tt I} (m)}}, \bm{x}_{1:L}; \bm{\theta}^{\tt II}) \nonumber
\end{align}

It is simple to compute $\triangledown_{\bm{\theta}^{\tt II}} \bar{\mathcal{L}}_{y}$. However, computing $\triangledown_{\bm{\theta}^{\tt I}} \bar{\mathcal{L}}_{y}$ is not trivial, because it requires differentiating through $\hat{\bm{y}}^{{\tt I} (m)}_{1:T^{{\tt I} (m)}}$, a random sample drawn from a distribution. The sampling process can be formulated as a deterministic function of $\bm{\theta}^{\tt I}$, using reparameterization tricks, such as Gumbel softmax \cite{jang2017categorical}. In general, suppose $\bm{y}^{\tt I}_{1:T^{\tt I}}$ can be viewed as a function of a random variable $\bm{z}$, and drawing samples from its distribution $p(\bm{z})$ is practical. 
Then instead of drawing $\{ \hat{\bm{y}}^{{\tt I} (1)}_{1:T^{{\tt I} (1)}}, ..., \hat{\bm{y}}^{{\tt I} (M)}_{1:T^{{\tt I} (M)}} \}$ from $p(\bm{y}^{\tt I}_{1:T^{\tt I}} | \bm{x}_{1:L}; \bm{\theta}^{\tt I})$, we can draw $\{ \hat{\bm{z}}^{(1)}, ..., \hat{\bm{z}}^{(M)} \}$ from $p(\bm{z})$:
\begin{align}
&\hat{\bm{z}}^{(m)} \sim p(\bm{z}); \;
\hat{\bm{y}}^{{\tt I} (m)}_{1:T^{{\tt I} (m)}} = f(\hat{\bm{z}}^{(m)}, \bm{x}_{1:L}; \bm{\theta}^{\tt I}) 
\end{align}
The gradients for $\bm{\theta}^{\tt I}$ and $\bm{\theta}^{\tt II}$ are
\begin{align}
& \triangledown_{\bm{\theta}^{\tt I}} \bar{\mathcal{L}}_{y}(\bm{\theta}^{\tt I}, \bm{\theta}^{\tt II})  \approx \label{eq:mp_loss_approx_grad1} \\
& \textstyle - \frac{1}{M} \sum_{m=1}^{M} ( \triangledown_{\hat{\bm{y}}^{{\tt I} (m)}_{1:T^{{\tt I} (m)}}} \log \hat{F}^{{\tt II} (m)} ) ( \triangledown_{\bm{\theta}^{\tt I}} \hat{\bm{y}}^{{\tt I} (m)}_{1:T^{{\tt I} (m)}} ) \nonumber\\
& \triangledown_{\bm{\theta}^{\tt II}} \bar{\mathcal{L}}_{y}(\bm{\theta}^{\tt I}, \bm{\theta}^{\tt II}) \approx \label{eq:mp_loss_approx_grad2}\\
& \textstyle - \frac{1}{M} \sum_{m=1}^{M} \triangledown_{\bm{\theta}^{\tt II}} \log \hat{F}^{{\tt II} (m)} \nonumber
\end{align}
Note that $\bm{y}^{\tt I}_{1:T^{\tt I}}$ is independent of $\bm{\theta}^{\tt I}$, and the summation in equation \ref{eq:mp_loss_naive} is over the entire space of $\bm{y}^{\tt I}_{1:T^{\tt I}}$. Hence $\triangledown_{\bm{\theta}^{\tt I}} p(\bm{y}_{1:T} | \bm{y}^{\tt I}_{1:T^{\tt I}}, \bm{x}_{1:L}; \bm{\theta}^{\tt II}) = 0$, and equations \ref{eq:mp_loss_grad1} and \ref{eq:mp_loss_grad1_approx} hold. In contrast, $\hat{\bm{y}}^{{\tt I} (m)}_{1:T^{{\tt I} (m)}}$ depends on $\bm{\theta}^{\tt I}$, and the summation in equation \ref{eq:mp_loss_approx} is over a set depending on $\bm{\theta}^{\tt I}$. $\hat{\bm{y}}^{{\tt I} (m)}_{1:T^{{\tt I} (m)}}$ can be viewed as a deterministic function of $\bm{\theta}^{\tt I}$. Hence $\triangledown_{\bm{\theta}^{\tt I}} p(\bm{y}_{1:T} | \hat{\bm{y}}^{{\tt I} (m)}_{1:T^{{\tt I} (m)}}, \bm{x}_{1:L}; \bm{\theta}^{\tt II}) \neq 0$, and equation \ref{eq:mp_loss_approx_grad1} holds.

For the second-pass model $\bm{\theta}^{\tt II}$, given the same group of samples, the gradients remain the same, i.e. $\triangledown_{\bm{\theta}^{\tt II}} \mathcal{L}_{y} = \triangledown_{\bm{\theta}^{\tt II}} \bar{\mathcal{L}}_{y}$, whether Monte Carlo approximation is applied to the loss or the gradients. For the first-pass model $\bm{\theta}^{\tt I}$, unless $M \to +\infty$, the gradients $\triangledown_{\bm{\theta}^{\tt I}} \mathcal{L}_{y}$ and $\triangledown_{\bm{\theta}^{\tt I}} \bar{\mathcal{L}}_{y}$ are usually different. It is not trivial to mathematically characterize the difference. However, from a practical point of view, it is simpler to apply Monte Carlo approximation to the gradients, which does not require any reparameterization trick. This explains why applying Monte Carlo approximation to the gradients is more common in existing research \cite{xia2017deliberation, hu2020deliberation}.




The joint training scheme has several drawbacks. 
As there is no loss over the intermediate output $\hat{\bm{y}}^{{\tt I} (m)}_{1:T^{{\tt I} (m)}}$, it is likely to deviate from valid target sequences. This makes it difficult to analyze the system. More importantly, if $\bm{\theta}^{\tt I}$ is randomly initialized, the intermediate output will be close to random noise, making it difficult for $\bm{\theta}^{\tt II}$ to learn to refine the intermediate output. In practice, it is common to pretrain $\bm{\theta}^{\tt I}$ with teacher forcing, in order to address these problems \cite{hu2020deliberation}.

In terms of efficiency, one important problem is that the sampling process is very often auto-regressive, in which case joint training cannot be run in parallel. Recently, Transformer-style models are widely used in various seq2seq tasks. One of their main advantages is parallel training. If teacher forcing is used, there is no recurrent connection in the model, and training can be done in parallel across the length $T$ of the output $\bm{y}_{1:T}$, because the reference output history $\bm{y}_{1:t-1}$ is available for any $t$. However, if sampling is required, these models must operate sequentially, because the generated output history $\hat{\bm{y}}_{1:t-1}$ must be used, which is not available beforehand.

\subsection{Separate Training} \label{sec:chDelib_training_separate}


For separate training, $\bm{\theta}^{\tt I}$ is trained as a standard sequence-to-sequence model with teacher forcing:
\begin{align}
\mathcal{L}_{y}(\bm{\theta}^{\tt I}) &= -\log p(\bm{y}_{1:T} | \bm{x}_{1:L}; \bm{\theta}^{\tt I})
\end{align}
Then it is fixed to generate samples $\{ \hat{\bm{y}}^{{\tt I} (1)}_{1:T^{{\tt I} (1)}}, ..., \hat{\bm{y}}^{{\tt I} (M)}_{1:T^{{\tt I} (M)}} \}$ for each input $\bm{x}_{1:L}$, using beam search or (a noisy version of) greedy search. Next, $\bm{\theta}^{\tt II}$ is again trained with teacher forcing:
\begin{align}
\mathcal{L}_{y}(\bm{\theta}^{\tt II}) &= -\log \textstyle \frac{1}{M} \sum_{m=1}^{M} \hat{F}^{{\tt II} (m)}
\end{align}
The gradients for $\bm{\theta}^{\tt I}$ and $\bm{\theta}^{\tt I}$ are
\begin{align}
&\triangledown_{\bm{\theta}^{\tt I}} \mathcal{L}_{y}(\bm{\theta}^{\tt I}) = \textstyle - \triangledown_{\bm{\theta}^{\tt I}} \log p(\bm{y}_{1:T} | \bm{x}_{1:L}; \bm{\theta}^{\tt I}) \label{eq:mp_loss_grad1_separate} \\
&\triangledown_{\bm{\theta}^{\tt II}} \mathcal{L}_{y}(\bm{\theta}^{\tt II}) = \textstyle - \frac{1}{M} \sum_{m=1}^{M} 
\triangledown_{\bm{\theta}^{\tt II}} \log \hat{F}^{{\tt II} (m)}
\label{eq:mp_loss_grad2_separate}
\end{align}
For $\bm{\theta}^{\tt II}$, the gradient is the same for separate training and joint training, given the same group of samples. This can be seen by comparing equations \ref{eq:mp_loss_grad2_approx} and \ref{eq:mp_loss_grad2_separate}. For $\bm{\theta}^{\tt I}$, however, there is a major difference. As described in the previous subsection, joint training uses a noisy Monte Carlo estimator for either the loss or the gradient, and is empirically found to be unstable \cite{xia2017deliberation}. In contrast, for separate training, $\bm{\theta}^{\tt I}$ is trained with teacher forcing, and is free from the above problem. This can be seen by comparing equation \ref{eq:mp_loss_grad1_separate} to equations \ref{eq:mp_loss_grad1_approx} and \ref{eq:mp_loss_approx_grad1}.

The separate training approach has the advantage that it allows parallel training. As $\bm{\theta}^{\tt I}$ is fixed, the sample $\hat{\bm{y}}^{\tt I}_{1:T^{\tt I}}$ can be generated and stored beforehand. So that when predicting $\bm{y}_{t}$, all the required information is available, including $\bm{x}_{1:L}$, $\bm{y}_{1:t-1}$ and $\hat{\bm{y}}^{\tt I}_{1:T^{\tt I}}$. 

\subsection{Discussion}

For all the approaches described above, there is a common choice: when updating the parameters of the second-pass model $\bm{\theta}^{\tt II}$, the first-pass model $\bm{\theta}^{\tt I}$ runs in free running mode. In other words, during training, the output $\hat{\bm{y}}^{{\tt I}}_{1:T^{{\tt I}}}$ from $\bm{\theta}^{\tt I}$ is generated in free running mode, instead of teacher forcing mode. This trains $\bm{\theta}^{\tt II}$ to fix the errors made by $\bm{\theta}^{\tt I}$ in free running mode, i.e. to address exposure bias. Empirically, it is shown that if the $\bm{\theta}^{\tt I}$ runs in teacher forcing mode while $\bm{\theta}^{\tt II}$ is trained, the network will not have any performance gain at the inference stage \cite{dou2021deliberation}.

Intuitively, when computing the distribution of $\bm{y}_{1:T}$, knowing $\hat{\bm{y}}^{{\tt I} }_{1:T^{{\tt I} }}$ leads to some information gain. This can be quantified by mutual information between $\bm{y}_{1:T}$ and $\hat{\bm{y}}^{{\tt I} }_{1:T^{{\tt I} }}$. Assuming that the expectation over the intermediate output is approximated by a single sample, the information gain can be formulated as:
\begin{align}
\mathcal{G}(\bm{y}_{1:T}, \hat{\bm{y}}^{{\tt I} }_{1:T^{{\tt I} }}) &= \textstyle \frac{1}{T} \sum_{t=1}^{T} \mathcal{H}( p(\bm{y}_{t} | \bm{y}_{<t}; \bm{\theta}^{\tt I}) ) -\label{eq:info_fr} \nonumber \\
&\mathcal{H}( p(\bm{y}_{t} | \bm{y}_{<t}, \hat{\bm{y}}^{{\tt I} }_{1:T^{{\tt I} }}; \bm{\theta}^{\tt I}, \bm{\theta}^{\tt II}) ) 
\end{align}
Here the input sequence is omitted for simplicity. $\mathcal{G}(\bm{y}_{1:T}, \hat{\bm{y}}^{{\tt I} }_{1:T^{{\tt I} }})$ denotes the information gain from condition the distribution of $\bm{y}_{1:T}$ on $\hat{\bm{y}}^{{\tt I} }_{1:T^{{\tt I} }}$. The intermediate output sequence is generated in free running mode, and is denoted $\hat{\bm{y}}^{{\tt I} }_{1:T^{{\tt I} }}$. Alternatively, the output sequence can be generated in teacher forcing mode. Let $\check{\bm{y}}^{{\tt I} }_{1:T^{{\tt I} }}$ the alternative sequence, the information gain becomes:
\begin{align}
\mathcal{G}(\bm{y}_{1:T}, \check{\bm{y}}^{{\tt I} }_{1:T^{{\tt I} }}) &= \textstyle \frac{1}{T} \sum_{t=1}^{T} \mathcal{H}( p(\bm{y}_{t} | \bm{y}_{<t}; \bm{\theta}^{\tt I}) ) -\label{eq:info_ft} \nonumber \\
&\mathcal{H}( p(\bm{y}_{t} | \bm{y}_{<t}, \check{\bm{y}}^{{\tt I} }_{1:T^{{\tt I} }}; \bm{\theta}^{\tt I}, \bm{\theta}^{\tt II}) ) 
\end{align}
We hypothesize that $\mathcal{G}(\bm{y}_{1:T}, \check{\bm{y}}^{{\tt I} }_{1:T^{{\tt I} }}) \leq \mathcal{G}(\bm{y}_{1:T}, \hat{\bm{y}}^{{\tt I} }_{1:T^{{\tt I} }})$, because the teacher forcing sequence is very similar to the reference sequence $\bm{y}_{1:T}$, and adds less information than the free running sequence. Experiments will be conducted in our future work to test this hypothesis.

So far in this section, it has been assumed that the goal of training is to learn a natural distribution. There are alternative options. For example, the entire network can be trained with MBR training, as described in section \ref{sec:intro_inf_train}. This is often adopted in tasks with a gold-standard objective metric, e.g. word error rate for ASR \cite{hu2020deliberation, hu2021transformer}. While this work focuses on supervised training, deliberation networks can also be used in unsupervised training. For example, a cycle consistency loss can be used in tasks where there is not a lot of paired data, such as image-to-image translation and voice conversion \cite{he2019deliberation}.

\section{Application Considerations} \label{sec:application}

When applying deliberation networks, it is essential to consider the nature of the input and the output. There are two key factors to consider. The first is whether the input and output have the same continuity. If so, they can be embedded in the same way. Examples include NMT and voice conversion. More precisely speaking, the initial input $\bm{x}_{1:L}$ and the first-pass output $\bm{y}^{\tt I}_{1:T^{\tt I}}$ are either both continuous or both discrete. Hence the additional encoder and attention mechanism for the first-pass output, $\bm{\theta}_{h,y}^{\tt II}$ and $\bm{\theta}_{\alpha,y}^{\tt II}$, can have exactly the same structure as those for the initial input, $\bm{\theta}_{h,x}^{\tt II}$ and $\bm{\theta}_{\alpha,x}^{\tt II}$. When the input and output are different in terms of continuity, the additional encoder needs to be modified. For example, for TTS, $\bm{x}_{1:L}$ is a discrete text sequence, and $\bm{y}^{\tt I}_{1:T^{\tt I}}$ is a continuous speech sequence.\footnote{
In most cases $\bm{y}^{\tt I}_{1:T^{\tt I}}$ is a feature sequence, and a neural vocoder maps it to a waveform.
}
Here the text embedding layer in $\bm{\theta}_{h,x}^{\tt II}$ can be replaced by a linear layer in $\bm{\theta}_{h,y}^{\tt II}$.
The second key factor to consider is whether the output is naturally discrete or continuous. The rest of this section will discuss both cases.

\subsection{Discrete output}

On a historical note, deliberation networks were first introduced for sequence-to-sequences tasks where both the input and output are text, such as NMT \cite{xia2017deliberation}. Their application was later extended to ASR, where the input is audio and the output is text. For these tasks, the additional attention connects two text sequences, which are naturally discrete. Compared with audio sequences, which are naturally continuous, text sequences are usually shorter and the tokens are less correlated in time. Therefore, it is easier for the additional attention to learn to align the sequences, and the standard training approaches in section \ref{sec:chDelib_training} work out-of-the-box \cite{hu2020deliberation, dou2022improving}.

When it comes to ASR, streaming is an increasingly important demand \cite{hu2020deliberation, hu2021transformer, Mavandadi2021Deliberation}. Typically, the first-pass model is a streaming model such as RNN-Transducer (RNN-T) \cite{he2019streaming}, and the RNN-T loss for the first-pass model is combined with the likelihood loss for the second-pass model, described in section \ref{sec:chDelib_training}. In some cases, MBR training is also applied to the second-pass model, directly optimizing the word error rate \cite{hu2020deliberation, hu2021transformer}. To improve the streaming outputs, the second-pass model often adopts more powerful building blocks, such as Transformer blocks. The most common training scheme is separate training followed by joint training. During joint training, the first-pass model generates samples sequentially, but this is less problematic than the other application cases, thanks to the efficiency of the first-pass model.

\subsection{Continuous output} \label{sec:TTS}

Continuous sequences, such as audio, are usually longer than discrete sequences, such as text. In general, longer sequences are harder for the attention mechanism, and reducing the time resolution alleviates the problem. For example, a pyramid encoder is often used in attention-based ASR models \cite{chan2015listen}. Alternatively, when applying deliberation networks to TTS, adjacent frames in the first-pass output $\bm{y}^{\tt I}_{1:T^{\tt I}}$ can be stacked in groups, forming a shorter sequence, before being fed into the encoder \cite{dou2021deliberation}.

Another challenge for continuous sequences is the strong correlation across time, i.e. among the tokens, which makes it hard to find the right tokens to focus on. For deliberation networks, the second-pass model has two sources of information: the initial input sequence $\bm{x}_{1:L}$ and the first-pass output $\bm{y}^{\tt I}_{1:T^{\tt I}}$. When $\bm{y}^{\tt I}_{1:T^{\tt I}}$ is continuous, the second-pass model is likely to ignore $\bm{y}^{\tt I}_{1:T^{\tt I}}$, as learning to attend to $\bm{x}_{1:L}$ can be enough for reaching a local optimum during training. In this case, the attention over $\bm{x}_{1:L}$ does not produce any meaningful alignment, and the deliberation network degrades into a standard single-pass model. To tackle this issue, the attention can be regularized. This is relatively simple when the attention is expected to be monotonic. For example, when applying deliberation networks to TTS, a guided attention loss \cite{tachibana2018efficiently} can be added:
\begin{equation} \label{eq:gal}
\begin{split}
&\mathcal{L}_{\alpha}(\bm{\theta}^{\tt II}) = \textstyle \sum_{t=1}^{T} \sum_{l=1}^{T^{\tt I}} [\alpha_{y,t,l} w_{t,l}]\\
&w_{t,l} = 1 - \exp{(-(t/T-l/T^{\tt I})/2g^{2})}
\end{split}
\end{equation}
where $g$ is a hyperparameter controlling the sharpness, and $\alpha_{y,t,l}$ is an element of the attention map $\bm{\alpha}_{y,1:T}$.\footnote{
In the subscripts of $\alpha_{y,t,l}$ and $\bm{\alpha}_{y,1:T}$, $_{y}$ indicates that the attention is over the first-pass output $\bm{y}^{\tt I}_{1:T^{\tt I}}$, and $_{t,l}$ is the position in the map.
} 
This encourages $\bm{\alpha}_{y,1:T}$ to be diagonal, enabling $\bm{\theta}^{\tt II}$ to make more extensive use of $\bm{y}^{\tt I}_{1:T^{\tt I}}$. For $\bm{\theta}^{\tt II}$, the complete loss is
\begin{equation} \label{eq:chMP_full_loss}
\mathcal{L}_{y,\alpha} (\bm{\theta}^{\tt II}) = \mathcal{L}_{y} (\bm{\theta}^{\tt II}) + \gamma \mathcal{L}_{\alpha} (\bm{\theta}^{\tt II})
\end{equation}
where $\gamma$ is a scaling factor. When $\mathcal{L}_{\alpha}$ is used, it is important to monitor $\bm{\alpha}_{y,1:T}$ and the inference performance on a validation set via objective metrics such as Global Variance \cite{dou2021deliberation}. When $\bm{\alpha}_{y,1:T}$ is sharply diagonal, $\mathcal{L}_{\alpha}$ is low, but the inference performance may degrade.


\section{Conclusion}
This paper introduces a unifying framework for deliberation networks, investigating various training options and application considerations. The key insights are as follows. First, to deal with the intractable marginalization of the intermediate output, it is simpler to apply Monte Carlo approximation to the gradients instead of the loss. Second, parallel training is possible for deliberation networks, as long as each pass is trained separately. Third, regardless of the application, it is essential that when training the parameters of a certain pass, its previous pass runs in free running mode. Finally, for applications where the output is continuous, a guided attention loss can be used to prevent the multi-pass model from degrading into a single-pass model.


\bibliography{custom}
\bibliographystyle{acl_natbib}

\end{document}